\documentclass[11pt,a4paper]{article}

\usepackage[a4paper,margin=2.5cm]{geometry} 
\usepackage[T1]{fontenc}
\usepackage[utf8]{inputenc}
\usepackage{lmodern}
\usepackage{setspace}
\usepackage{amsmath,amssymb,amsfonts,amsthm}
\usepackage{mathrsfs}
\usepackage{graphicx}
\usepackage{multirow}
\usepackage{booktabs}
\usepackage{algorithm}
\usepackage{algpseudocode}
\usepackage{listings}
\usepackage{xcolor}
\usepackage{url}
\usepackage{subfig}
\usepackage{rotating}
\usepackage{appendix}
\usepackage[authoryear]{natbib}
\usepackage{pifont}
\usepackage[hidelinks]{hyperref}

\title{Generating Multiple-Choice Knowledge Questions with Interpretable Difficulty Estimation using Knowledge Graphs and Large Language Models}

\author{
Mehmet Can Şakiroğlu$^{1}$ \and
H. Altay Güvenir$^{1}$ \and
Kamer Kaya$^{2,3}$\thanks{Corresponding author: \texttt{kaya@sabanciuniv.edu}}
}

\date{}

\begin{document}

\maketitle

\begin{center}
$^{1}$ Computer Engineering Department, Bilkent University, Ankara, Turkey\\
$^{2}$ Faculty of Engineering and Natural Sciences, Sabanci University\\
$^{3}$ VERIM, Center of Excellence in Data Analytics, Sabanci University
\end{center}

\vspace{1em}

\begin{abstract}
Generating multiple-choice questions (MCQs) with difficulty estimation remains challenging in automated MCQ-generation systems used in adaptive, AI-assisted education. This study proposes a novel methodology for generating MCQs with difficulty estimation from the input documents by utilizing knowledge graphs (KGs) and large language models (LLMs). Our approach uses an LLM to construct a KG from input documents, from which MCQs are then systematically generated. Each MCQ is generated by selecting a node from the KG as the key, sampling a related triple or quintuple---optionally augmented with an extra triple---and prompting an LLM to generate a corresponding stem from these graph components. Distractors are then selected from the KG. For each MCQ, nine difficulty signals are computed and combined into a unified difficulty score using a data-driven approach. Experimental results demonstrate that our method generates high-quality MCQs whose difficulty estimation is interpretable and aligns with human perceptions. Our approach improves automated MCQ generation by integrating structured knowledge representations with LLMs and a data-driven difficulty estimation model.
\end{abstract}


\section{Introduction}
\label{section1}

\begin{figure}[htb]
    \centering
    \includegraphics[width=0.67\linewidth]{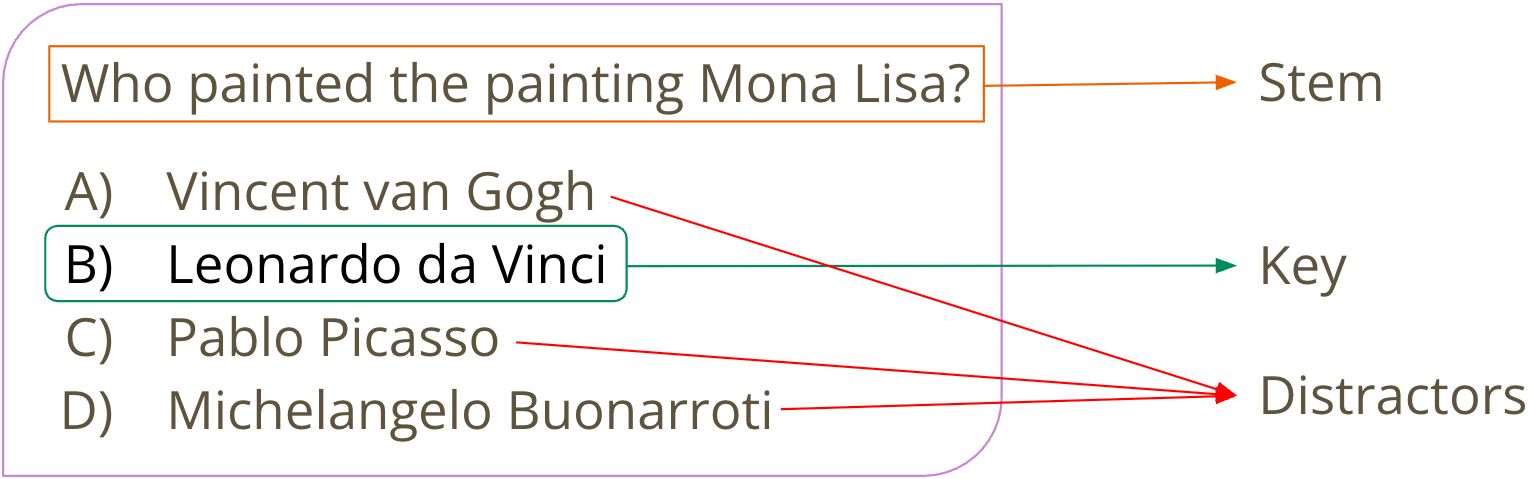}
    \caption{An example multiple-choice question.}
    \label{fig:mcq}
\end{figure}

Multiple-choice questions are a common form of assessment used in educational settings due to their utility in standardized testing, self-assessment, and adaptive learning systems. A multiple-choice question typically consists of a question or statement known as the ``{\em{stem}}" followed by several possible answers. One of these answers is the correct one, referred to as the ``{\em key}", while the others are incorrect, known as ``{\em distractors}". An example multiple-choice question whose parts are annotated can be seen in Figure~\ref{fig:mcq}. The effectiveness of a multiple-choice question depends on the quality of its stem, key, and distractors, as they must be clear, concise, and appropriately challenging for the intended audience. Automated multiple-choice question generation offers the potential to reduce the time and effort required for manual question generation in education while maintaining quality and relevance.

Even though multiple-choice questions are well-known and widely used, automatically generating relevant, high-quality, and diverse multiple-choice knowledge questions with realistic difficulty estimation remains a challenging task.
Existing methods either rely on handcrafted rules, which lack scalability, or black-box deep-learning approaches, which struggle with interpretability. An effective MCQ-generation system must not only generate high-quality questions but also provide an interpretable difficulty estimation aligned with human perception.
Furthermore, another challenge lies in the cost of creating an appropriate dataset, as it requires extensive human annotation.

\begin{figure}[htb]
    \centering
    \includegraphics[width=0.83\linewidth]{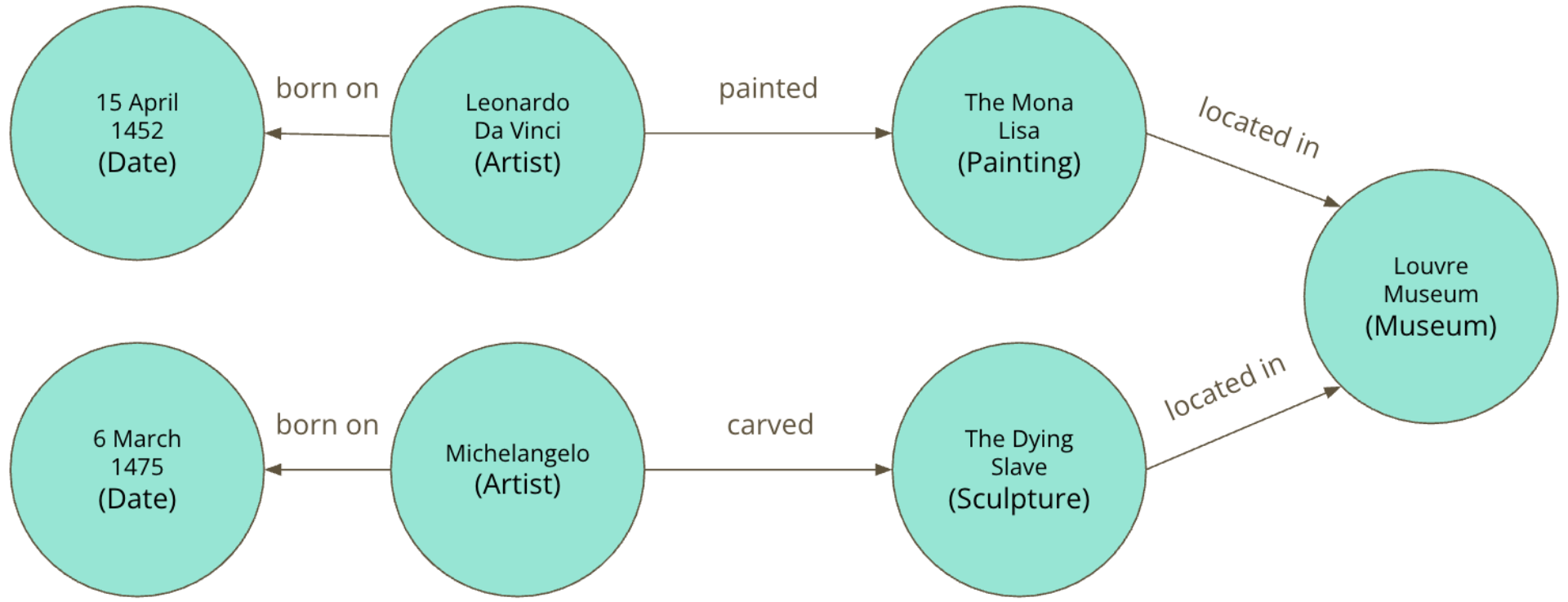}
    \caption{An illustrative example of a knowledge graph representing entities, their types, and the semantic relationships between them.}
    \label{fig:kg}
\end{figure}

To address such challenges, we propose a novel approach that combines knowledge graphs (KGs) with large language models (LLMs) for generating multiple-choice questions with interpretable difficulty estimation. The integration of LLMs enables advanced natural language understanding and generation capabilities, while the knowledge graph serves as a structured backbone for representing factual information and entity relationships. An illustrative example with a small knowledge graph is presented in Figure~\ref{fig:kg}, where entities, their types, and semantic relationships are encoded as nodes and edges.

Our method begins by building a knowledge graph (KG) from the input text using an LLM. Multiple-choice questions (MCQs) are then systematically generated from this graph. For each MCQ, a key node serving as the correct answer is selected, along with a relevant triple or quintuple from the KG, which may be supplemented with an extra triple for additional context. These structured graph elements are then converted into a question stem by the LLM, and suitable distractors are also selected from the KG. To ensure explainability in difficulty estimation, we compute several intuitive and interpretable difficulty signals, including graph- and network-analytics-based metrics, embedding-based semantic similarities, and linguistic features, and combine them into a unified difficulty score in a data-driven manner by leveraging the human evaluation data we collect. Furthermore, beyond its technical aspects, our framework has substantial potential for enhancing educational practices, as it can be integrated into intelligent tutoring systems, adaptive testing platforms, and digital learning environments to deliver personalized assessments. Overall, the novel contributions of this work are summarized below:

\begin{itemize}
\item We propose a novel end-to-end framework for generating multiple-choice questions from textual sources. The framework constructs a knowledge graph from the input text, generates MCQs from the resulting graph, and estimates their difficulty in an interpretable manner. Our hybrid approach combines the structural knowledge of KGs with the natural language generation capabilities of LLMs to produce high-quality MCQs.
\item We construct a novel dataset comprising MCQs, their corresponding knowledge subgraphs, and empirically derived difficulty labels based on an average of 38 independent human responses per question. To the best of our knowledge, this is the first dataset that jointly includes MCQs, their underlying knowledge subgraphs, and empirical difficulty labels derived from broad human evaluation.
\item We introduce nine interpretable, data-driven difficulty estimation signals spanning graph topology, textual semantics, and linguistic complexity, thereby capturing both structural and semantic properties of the generated MCQs. Their effectiveness is assessed through four regression models and an ablation study.
\item Our data and code are available to enable reproduction of this study and further research.\footnote{\url{https://github.com/kamerkaya/Generating-Multiple-Choice-Knowledge-Questions-from-KGs}}
\end{itemize}

The remainder of this paper is organized as follows: Section~\ref{section2} discusses related work in question generation, MCQ generation, difficulty modeling, and the usage of knowledge graphs and LLMs in these settings. Section~\ref{section3} details our methodology, including knowledge graph construction, multiple-choice question generation, and difficulty estimation. Section~\ref{section4} details the dataset construction process, including the human evaluation protocol used to obtain empirical difficulty labels, and presents key statistics characterizing the resulting dataset. Section~\ref{section5} presents the experimental results and analysis, and Section~\ref{section6} discusses them. Finally, Section~\ref{section7} concludes the paper, points out key limitations, and outlines directions for future research accordingly.

\section{Background and Related Work}
\label{section2}

Automated question generation has seen significant advancements with various approaches exploring different aspects of the problem.
A review of related works on this topic, including their improvement aspects and limitations, is conducted with comparisons to our own. 

\subsection{Graph- and Ontology-based Question Generation}

\citet{17sathish} propose a method that employs recurrent neural networks (RNNs) rather than templates, as in their prior work, for question verbalization, thereby enhancing robustness and reducing the need for manual effort. However, the authors do not discuss the difficulty level of the question. Instead, they directly focus on the question text (stem) generation with the correct answer (key).

\citet{18elsahar} present an encoder-decoder architecture that leverages textual contexts and a copy mechanism to address the challenge of the inability to generate questions for unseen predicates and entity types of previous works, hence enabling zero-shot question generation. Yet, Elsahar et al. restrict their task to generating natural language questions given a single input triple instead of multiple triples. They also do not emphasize parameters such as difficulty.

Later, \citet{23chen} propose a bidirectional Graph2Seq (Graph-to-Sequence) model to encode the KG subgraph to take advantage of the information provided by the graph structure.
Then, they use an RNN decoder with a node-level copying mechanism to generate the question based on the output of a GNN-based (Graph Neural Network) encoder they propose. However, Chen et al. do not control or investigate the difficulty of the generated questions.

\citet{23li} take advantage of a pre-trained LLM, GPT-2, to obtain a comprehensive semantic context, and then they construct a graph using the entities identified within this context. Then, they utilize an answer-aware graph attention network (GAT) to update it based on the constructed graph to generate the question. However, the constructed graph is not a KG and its edges do not represent semantic relationships; instead, they represent the co-occurrence of two entities in the same paragraph. The authors also do not discuss the difficulty of the generated question.

\subsubsection{Approaches Taking Difficulty into Account}

\citet{16alsubait} generate difficulty-controllable multiple-choice questions by utilizing ontologies to measure the similarity between the distractors and the key, based on the number of common properties they share in an ontology. They handle the stem generation with a template-based approach based on the ontology.

\citet{16seyler} provide a KG-based question generation technique that first selects an entity from the KG and then generates a SPARQL (SPARQL Protocol and RDF Query Language) query that specifies that node uniquely and converts the query to a natural language question with preset templates referred to as {\em question verbalization}. For difficulty estimation, their approach requires a question-answer corpus with annotated difficulties to train a classifier on. They propose generating distractors by relaxing their original query's constraints to retrieve more than one unique answer. The effect of the distractors on the question's difficulty is defined with the {\em confusion} metric computed via the difficulty classifier trained on the Q\&A corpus.

\citet{19kumar} study on generating complex, multi-hop questions that require reasoning across multiple KG triples, hence accepting subgraphs and multiple triples as its input, as opposed to a single triple. They present a Transformer-based model that also takes the difficulty into account. This work is similar to ours as it shares the goal of enabling the generation of multi-hop questions with controllable difficulty levels. However, Kumar et al. define difficulty exclusively based on two NER-dependent factors: (1) confidence scores for entity-mention linking and (2) the selectivity of entity surface forms, without incorporating structural or semantic properties of the question or the subgraph as a whole.

\citet{kusuma2022} propose an ontology-based question generation framework that introduces a combination of taxonomy ontology and sentence ontology, referred to as knowledge ontology. Given a textual input, their system constructs an ontology and generates various types of questions using query templates. Both the type and difficulty of each question are determined manually by ontology engineers and domain experts. While their approach supports multiple question formats and provides difficulty annotations, it relies on template-based generation and does not incorporate automated difficulty estimation.

\citet{bi24} introduce DiffQG, a difficulty-controllable single-answer question generation model with no distractors that produces natural language questions from a given KG subgraph with a specified difficulty level.
Their approach uses a mixture-of-experts module to learn soft templates for different difficulty levels, enhancing the diversity of question phrasing without relying on manually crafted templates.
They also incorporate a disentanglement module that isolates the KG triples relevant to the target difficulty, which enables counterfactual reasoning—training the model on perturbed subgraphs to reinforce a causal link between the difficulty label and the question features.
Additionally, the authors propose a difficulty estimation mechanism called AutoDE that considers eight difficulty signals and the use of those signals by normalizing and linearly combining them. We follow a similar approach to estimate the difficulty with different signals, yet the assumption of linearity can be misleading, and to solve that, we propose a data-driven methodology while employing both linear and non-linear models.
Moreover, they propose the counterfactual reasoning approach to further enable the generation of questions of different difficulty levels given the same input subgraph, with the underlying idea that the given subgraph should not dictate the difficulty of the question. On the contrary, we think the topology of the given subgraph and the information within should be the main determining factor for difficulty, especially in knowledge questions.
The broader challenge of text-based question difficulty prediction is systematically reviewed by \citet{AlKhuzaey2023}, who highlight the major role of linguistic features and the need for standardized datasets to enable meaningful comparisons between models.

\citet{zhu24} quantify difficulty as the average similarity of each distractor to the correct answer. In their system, an MCQ can be generated in multiple versions: by selecting distractors that are very similar to the answer, the question’s difficulty score increases, whereas using more dissimilar distractors yields an easier question. They implement an algorithm to automatically pick a set of distractors such that the resulting question’s difficulty falls into a specified range or level. If impossible, they automatically lower the targeted level by one. To improve the accuracy of this difficulty metric, the authors refine the similarity calculation with topological weights, effectively incorporating knowledge from the structure of a semantic network to better judge how “close” a distractor is to the answer. This approach allows for explicit difficulty control: the same question stem can be paired with easier or harder sets of options, producing tiered versions of the question. One limitation is that this method predominantly captures only one aspect of difficulty: distractor ambiguity. It does not account for other factors like the intrinsic complexity of the question’s content. Additionally, while the difficulty calculation is interpretable, they do not investigate a comprehensive list of signals that might affect the difficulty of an MCQ.

\citet{wei24} introduce KGNN-ADP, a KG-enhanced neural network designed for predicting the difficulty of an MCQ. Their framework leverages a domain-specific KG to extract relevant knowledge points for a given question and assess difficulty along two primary axes: (1) knowledge difficulty, computed by analyzing the semantic proximity and structural weight of the matched knowledge points in the graph, and (2) option difficulty, modeled using both the semantic similarity between distractors and the answer and the internal semantic divergence within the full question. While this work shares our emphasis on difficulty estimation grounded in KGs, it differs from ours in scope: their model predicts difficulty for existing MCQs rather than generating new ones by solely focusing on the absolute difficulty of MCQs, and not on the broader task of MCQ generation from raw text or graphs.

The prior research also extends to specialized topics. For example, an ontology-based approach by \citet{Leo2019} has been applied in the medical field to generate multi-term, case-based multiple-choice questions (MCQs) from a medical ontology, which shows the value of structured knowledge in niche areas. Also, a hybrid framework by \citet{Kumar2023} combines machine-learning and semantic techniques to create different types of MCQ stems for technical fields, which are then assessed against the cognitive levels of Bloom's Taxonomy.

Additionally, the synergy between knowledge graphs and large language models has recently gained attention for related tasks as well, such as question answering, where combining structured KGs with LLMs improves performance \citep{huang2025, jiang2024}.

\subsection{Non-graph-based Approaches}

\citet{vachev22} present {\em Leaf}, a system for the end-to-end generation of MCQs from educational texts. The goal is to ease quiz/exam creation for instructors by inputting course material and outputting factual MCQs. The system uses neural models to perform the sub-tasks of question and distractor generation: it fine-tunes a Transformer-based text-to-text model (T5) on Q\&A pairs and trains another model to produce realistic distractors. Given a passage and a target answer, the former produces a question that becomes the input of the latter.
This system demonstrates that large-scale pre-trained models can be harnessed to automate MCQ creation from unstructured text, achieving good fluency and relevance in both questions and distractors. {\em Leaf}, however, does not incorporate question difficulty; all the questions are intended to be high-quality, but there is no control on or estimation of the difficulty. Furthermore, {\em Leaf} relies on raw text input and does not utilize structured KGs; as such, it may struggle with the coverage of specific relationships or while providing explanations. On the contrary, our work aims to estimate and explain the difficulty of each question generated.

\citet{zeng2024} propose RTRL, a deep question generation framework that combines a relation-aware transformer with reinforcement learning for deep question generation. RTRL introduces an encoder that embeds answer distances and words, and uses a BiLSTM to build answer-contextualized representations. RTRL also models both explicit linguistic relations and implicit semantic relations via its relation-aware transformer. The model is fine-tuned using reinforcement learning with sentence-level rewards to optimize evaluation metrics, thereby addressing training-evaluation mismatch. Although RTRL significantly improves performance in deep question generation, it does not focus on multiple-choice question generation or difficulty modeling; its focus remains on deep question generation.

The research in this field has also moved beyond English-language applications, with works like \citet{Johnson2024} demonstrating a parallel construction method to generate questions for Spanish textbooks by leveraging an existing English-based system.

\subsubsection{Approaches Taking Difficulty into Account}

\citet{gao19} explores difficulty control while generating open-ended questions for reading comprehension. They use text passages, where the input is a sentence from a reading comprehension article plus a target answer phrase, and the output is a question that asks about that answer at a specified difficulty level. A dataset with difficulty tags (over 70k questions split into easy and hard), enabling the authors to train an end-to-end sequence-to-sequence question generation model with difficulty, is leveraged. The results showed that their model can indeed tailor the generated questions to the requested difficulty without losing quality, as the questions remain fluent and answerable while matching the difficulty specification. Gao et al. thereby demonstrate the feasibility of controlling question difficulty. However, their work is limited to open-ended Q\&A pairs in text and a binary difficulty distinction, since there is no provision for generating distractors as in MCQs.

\citet{cheng21} advances \citet{gao19}'s approach by focusing on the reasoning complexity of questions. Noting that the prior method offered little interpretability on difficulty, they redefine the difficulty as the number of inference hops or reasoning steps required to answer it. This strategy offers better interpretability and stronger logical controllability than treating difficulty as a latent label. However, the approach is applied to free-text Q\&A settings and generates single-answer questions; it does not handle MCQs. Moreover, it restricts the definition of difficulty to only a single parameter: the number of reasoning steps required.

\subsection{A Brief Comparison with the Literature}

Our work distinguishes itself from the literature by integrating easy-to-achieve structured knowledge with auto-generated knowledge graphs and large language models to generate high-quality MCQs with interpretable difficulty estimation. Unlike prior QG work, we leverage a knowledge graph with an LLM, using the KG to ground the question in verifiable facts and the LLM to produce fluent questions by taking advantage of its pre-trained capabilities. This combination enables us to cover complex knowledge that might require reasoning while maintaining natural language fluency and diversity by leveraging LLMs, a synergy not explored in previous systems to the best of our knowledge. Furthermore, our difficulty estimation system provides justifications for a question’s difficulty rating based on the defined difficulty signals, offering interpretability. Lastly, we explicitly calibrate our difficulty predictions with the dataset we collect from human participants, ensuring that the difficulty levels correspond to the real world. This human-in-the-loop calibration means that our system’s idea of difficulty is not just an assumption but reflects actual ease or struggle observed among the participants. Through this novel integration of KGs, LLMs, and a human-calibrated approach, our work delivers a more comprehensive solution for generating multiple-choice questions and estimating their difficulty. A comparison of the literature above, including this work, is presented in Table~\ref{tab:related-work-comparison}.

\begin{table}[htbp]
  \caption{Comparison of the presented literature, including the proposed study.}
  \label{tab:related-work-comparison}
  \centering
  \footnotesize
    \setlength{\tabcolsep}{4pt}
  \begin{tabular}{@{}p{0.11\linewidth}|p{0.15\linewidth}|p{0.23\linewidth}|p{0.02\linewidth}|p{0.02\linewidth}|p{0.28\linewidth}|p{0.07\linewidth}@{}}
    \toprule
& \rotatebox{90}{\textbf{Input}} 
& \rotatebox{90}{\textbf{QG Model}} 
& \rotatebox{90}{\textbf{Mul.-hop QG}} 
& \rotatebox{90}{\textbf{MCQ Gen.}} 
& \rotatebox{90}{\parbox{1.5cm}{\textbf{Difficulty Modeling}} }
& \rotatebox{90}{\parbox{1.5cm}{\textbf{Human Calib. in Diff. Mod.}}} \\    \midrule\midrule
\cite{17sathish} & KG & RNN-based (seq2seq) & \ding{55} & \ding{55} & - & - \\
    \hline
\cite{18elsahar} & KG triple and related texts & Encoder-decoder with copy mechanism & \ding{55} & \ding{55} & - & - \\
    \hline
        \cite{vachev22} & Text & Transformer-based (T5) & \ding{55} &  \checkmark  & - & - \\
    \hline
\cite{23chen} & KG & Bidirectional graph2seq model with copy mechanism & \checkmark  & \ding{55} & - & - \\
    \hline
\cite{23li} & Text/auto-generated entity co-occurrence graph & GPT-2, answer-aware GAT, and multi-head attention generation module & \checkmark  & \ding{55} & - & - \\
    \hline
\cite{zeng2024} & Text & Relation-aware transformer with reinforcement learning & \checkmark  & \ding{55} & - & - \\
    \hline\hline
\cite{16alsubait} & Ontology & Template-based & \ding{55} & \checkmark  & Difficulty control via key/distractor similarities based on number of common properties & \ding{55} \\
    \hline
\cite{16seyler} & KG & SPARQL queries and templates & \ding{55} & \checkmark  & Difficulty estimation with logistic regression & \checkmark  (Ind.) \\
    \hline
\cite{19kumar} & KG & Transformer-based & \checkmark  & \ding{55} & Difficulty control and estimation with custom formula & \ding{55} \\
    \hline
\cite{gao19} & Text & Seq2seq encoder-decoder based on LSTMs with copy mechanism & \ding{55} & \ding{55} & Difficulty control by initializing the hidden state of the decoder w.r.t. the required difficulty & \checkmark  (Ind.) \\
    \hline
\cite{kusuma2022} & Text, and its auto constructed ontology & SPARQL queries and templates & \checkmark  & \ding{55} & Manual via ontology engineers \& domain experts & \checkmark  \\
    \hline
\cite{bi24} & KG & Soft-templates with MoE, and counterfactual reasoning & \checkmark  & \ding{55} & Difficulty control and estimation via linear modeling of proposed signals & \checkmark  \\
    \hline
\cite{zhu24} & Semantic Network & Rule-based & \ding{55} & \checkmark  & Difficulty estimation and discrete control based on key/distractor similarities by also integrating topological weights & \ding{55} \\
    \hline
\cite{wei24} & Existing MCQ and KG & - & - & - & Difficulty prediction with the proposed KG-enhanced neural network based on Bi-LSTM & \ding{55} \\
    \midrule\midrule
\textbf{This work} & Text/auto-gen. KG & Knowledge graph and LLM-based & \checkmark  & \checkmark  & Difficulty estimation with both linear and non-linear models based on signals & \checkmark  \\
    \bottomrule
  \end{tabular}
\end{table}

\section{Materials and Methods}
\label{section3}

\begin{figure}[htb]
    \centering
    \includegraphics[width=0.9\linewidth]{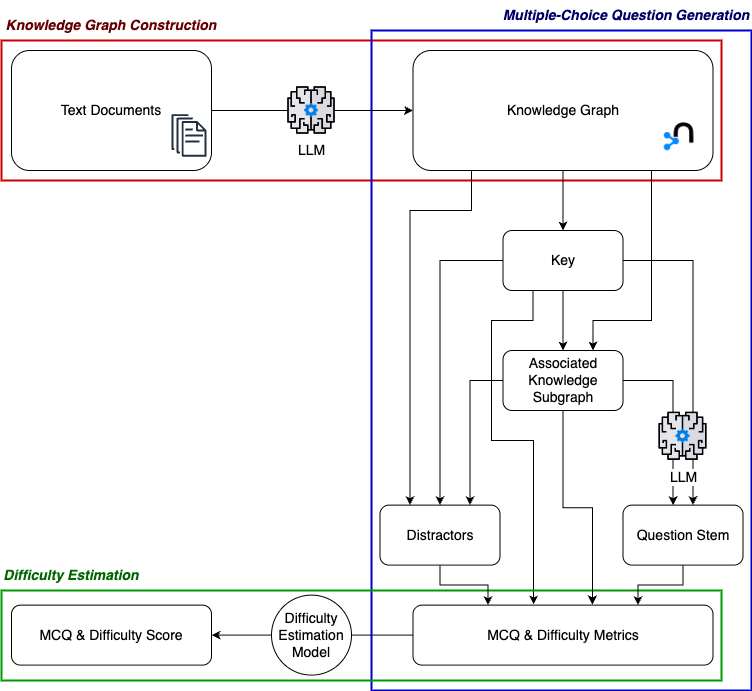}
    \caption{Overview of the proposed framework.}
    \label{fig:diagram}
\end{figure}

In this section, we present the methodology adopted in our study, which consists of three main components: KG construction, MCQ generation, and difficulty estimation. Figure~\ref{fig:diagram} provides an overview of the proposed framework. Given a corpus of factual information, we first construct a structured knowledge graph that captures the entities and their semantic relationships. This graph serves as the foundation for generating candidate MCQs by selecting appropriate subgraphs and transforming them into MCQs. Finally, we estimate the difficulty of each MCQ using a combination of graph-based, embedding-based semantic similarities, and linguistic features. Each stage of the framework is elaborated in the following subsections.

\subsection{Knowledge Graph Construction}

Given a set of input text documents, we construct a knowledge graph (KG) to serve as the foundation for multiple-choice question generation. To extract structured information from given documents, we apply a large language model (LLM) to identify salient factual statements and convert them into subject–predicate–object triples. These triples capture entity–relation–entity structures that represent core knowledge units within the text. The procedure is summarized in Algorithm~\ref{alg:kgc}, which outlines how each document is processed into a graph document and subsequently integrated into a unified KG stored in a Neo4j\footnote{\url{https://neo4j.com}} database.

\begin{figure*}[htb]
\centering
\begin{minipage}{0.92\textwidth}
\begin{algorithm}[H]
\caption{Knowledge Graph Construction}
\label{alg:kgc}
\begin{algorithmic}[1]
    \State \textbf{Input:} Set of documents $\mathcal{D} = \{doc_1, doc_2, \dots, doc_n\}$
    \For{each $doc_i \in \mathcal{D}$}
        \State $graphDocument_i \gets \texttt{LLMGraphTransformer}(doc_i)$ 
        \Comment{Extract nodes with types and relationships using an LLM, specifically with LangChain's LLMGraphTransformer}
    \EndFor
    \State $\mathcal{G} \gets \{graphDocument_1, \dots, graphDocument_n\}$ 
    \Comment{Aggregate all documents}
    \State \texttt{constructAndSaveKG}($\mathcal{G}$) 
    \Comment{Parse and persist the graph in Neo4j}
\end{algorithmic}
\end{algorithm}
\end{minipage}
\caption{The overview of the knowledge graph construction process from input documents using an LLM and integration into a Neo4j graph database.}
\end{figure*}

We begin by constructing the knowledge graph (KG) from a curated collection of textual documents. In this work, we use Wikipedia’s top-100 most-viewed articles\footnote{\url{https://en.wikipedia.org/wiki/Wikipedia:Popular_pages\#Top-100_list}}, which we automatically retrieved using a custom scraping script. The raw content is processed to extract factual knowledge using GPT-4o\footnote{\url{https://platform.openai.com/docs/models/gpt-4o}}, a large language model provided by OpenAI.

To transform unstructured text into structured graph representations, we employ \textsf{LLMGraphTransformer} from LangChain\footnote{\url{https://www.langchain.com}}. This tool ingests a list of documents and using the underlying LLM, it extracts subject–predicate–object triples to represent key factual relationships. These triples are subsequently stored in a Neo4j graph database. Thus, the KG captures both the semantic entities, their types, and the relations between them, which are later leveraged to generate MCQs.

The KG construction pipeline is designed to be schema-free, accommodating a wide variety of relations and entity types extracted from natural language. To facilitate downstream tasks, we also compute a node centrality metric, specifically degree centrality, and node embeddings for each entity in the graph using FastRP by \citet{fastrp}, which later serve as useful components for some of the difficulty signals. The entire construction process is automated, resulting in a rich and semantically meaningful representation of the source content.

\subsection{Multiple-Choice Question Generation}

Once the knowledge graph is constructed, we systematically generate MCQs by selecting high-centrality nodes as key candidates. Specifically, in this work, we choose the top-40 most central nodes in the KG based on degree centrality and attempt to generate four MCQs per node. The process that defines the generation of a single MCQ is outlined in Algorithm~\ref{alg:mcqgen}.

\begin{figure*}[htb]
\centering
\begin{minipage}{0.92\textwidth}
\begin{algorithm}[H]
\caption{Multiple-Choice Question Generation}
\label{alg:mcqgen}
\begin{algorithmic}[1]
    \State \textbf{Input:} \texttt{kg} \Comment{Knowledge graph constructed in Algorithm~\ref{alg:kgc}}
    \State \textbf{Input:} \texttt{keyNode} \Comment{Selected node to serve as the correct answer}
    \State \texttt{tripleOrQuintuple} $\gets$ Sample a triple or quintuple of \texttt{keyNode}
    \State \texttt{useExtraTriple} $\gets$ Set to \texttt{True} or \texttt{False}
    \If{\texttt{useExtraTriple}}
        \State \texttt{extraTriple} $\gets$ Retrieve another triple of \texttt{keyNode}
        \State \texttt{associatedSubgraph} $\gets$ \{\texttt{tripleOrQuintuple}, \texttt{extraTriple}\}
    \Else
        \State \texttt{associatedSubgraph} $\gets$ \{\texttt{tripleOrQuintuple}\}
    \EndIf
    \State \texttt{prompt} $\gets$ \texttt{constructPrompt(keyNode, associatedSubgraph)}
    \State \texttt{questionStem} $\gets$ \texttt{LLM(prompt)}
    \State \texttt{maxDepth} $\gets 5$ \Comment{Maximum BFS depth for distractor search}
    \State \texttt{distractors} $\gets$ \texttt{generateDistractors(kg, keyNode,}
    \Statex \hspace{7em} \texttt{associatedSubgraph, maxDepth)}
    \If{\texttt{validate(questionStem, keyNode, distractors)}}
        \State \texttt{difficultySignals} $\gets$ Compute the defined difficulty signals
        \State mcq $\gets$ \texttt{MCQ(questionStem, keyNode, distractors)}
        \State \texttt{save(mcq, difficultySignals)}
    \EndIf
\end{algorithmic}
\end{algorithm}
\end{minipage}
\caption{The overview of the MCQ generation process: The pipeline samples an associated subgraph from the KG given the key, prompts an LLM to generate a question stem from it, retrieves proper graph-based distractors, computes difficulty signals of the generated MCQ after validation, and saves the resulting MCQ with its computed difficulty signals.}
\end{figure*}

\begin{figure}[htb]
\centering
\includegraphics[width=0.95\linewidth]{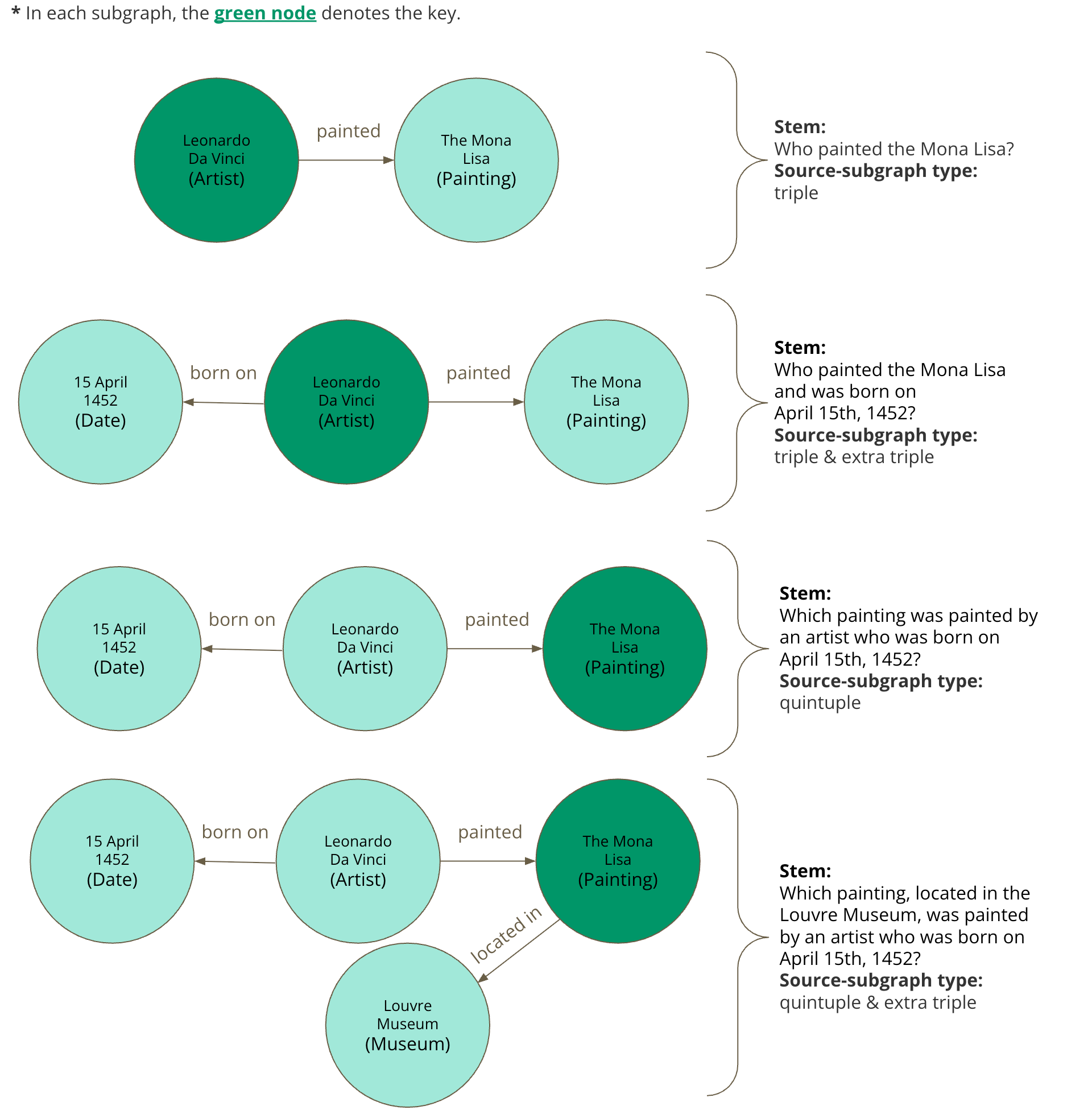}
\caption{Illustrative examples of each subgraph type alongside potential question stems that could be generated from them. The green node represents the key in each case.}
\label{fig:subgraphtypes}
\end{figure}

Each MCQ is generated from either a triple (single-hop question), a triple with additional context from an extra triple, a quintuple (double-hop question), or a quintuple with additional context from an extra triple. For each selected node (key), a related triple or quintuple is selected from the KG. Optionally, an extra triple may be added to provide additional context. These structured elements are passed to a prompt-based question generation module powered by GPT-4o to create the question stem. The prompts are crafted to follow the style of well-crafted trivia questions, inspired by formats such as those used in the quiz show \textit{Who Wants to Be a Millionaire?}. Figure~\ref{fig:subgraphtypes} presents illustrative examples of each subgraph type alongside potential question stems that could be generated from them.

Distractors are selected from the KG using a breadth-first search (BFS) strategy, ensuring they belong to the same semantic type as the key and are progressively distant in graph depth. In rare cases, the KG does not contain a sufficient number of appropriate distractors for a given key. In such instances, the MCQ generation attempt is aborted to preserve the quality of the generated MCQ. In our experiments, this limitation accounts for a total of 156 finalized MCQs instead of the originally targeted 160.

At this stage, having obtained the key, the question stem, and the distractors, the construction of the MCQ is complete. To ensure the quality and correctness of the generated MCQ, we perform a post-processing step for validation using a large language model. Each distractor option is independently evaluated by prompting the LLM with the question stem and the respective distractor. If the model identifies any distractor as a potentially correct answer, the MCQ is marked as invalid. This automated validation process serves as a safeguard against ambiguities or inconsistencies in the question formulation, ensuring that each MCQ has a single, unambiguous correct answer. Only those questions that pass this validation step are retained for subsequent analysis.

In cases where an MCQ fails the validation step, a retry mechanism is triggered to attempt regeneration. The process revisits the same key entity and initiates a new question generation sequence, potentially selecting a different supporting triple, quintuple, or extra triple, and generating a new set of distractors. This mechanism is repeated up to a predefined number of retries to ensure robustness and increase the likelihood of producing a valid MCQ. If all attempts fail, the MCQ is discarded from the final dataset to maintain overall quality.

Finally, each MCQ is saved with a set of metadata features referred to as difficulty signals, which are later utilized for difficulty estimation. All signals are normalized into the range \([0,1]\) using min-max normalization to ensure consistency and comparability across different feature types. These signals serve as interpretable proxies for MCQ difficulty and are further investigated and explained in Section~\ref{diffest}, where we also describe how they are integrated for a unified difficulty estimation model.

\subsection{Difficulty Estimation}
\label{diffest}

To estimate the difficulty of each generated MCQ, we extract and analyze a set of nine interpretable signals, each capturing a distinct aspect of the question's structure, semantics, or linguistic complexity. These signals detailed here serve both as standalone indicators and as input features to a regression model for predicting difficulty scores.

\begin{figure}[htb]
\centering
\includegraphics[width=0.93\linewidth]{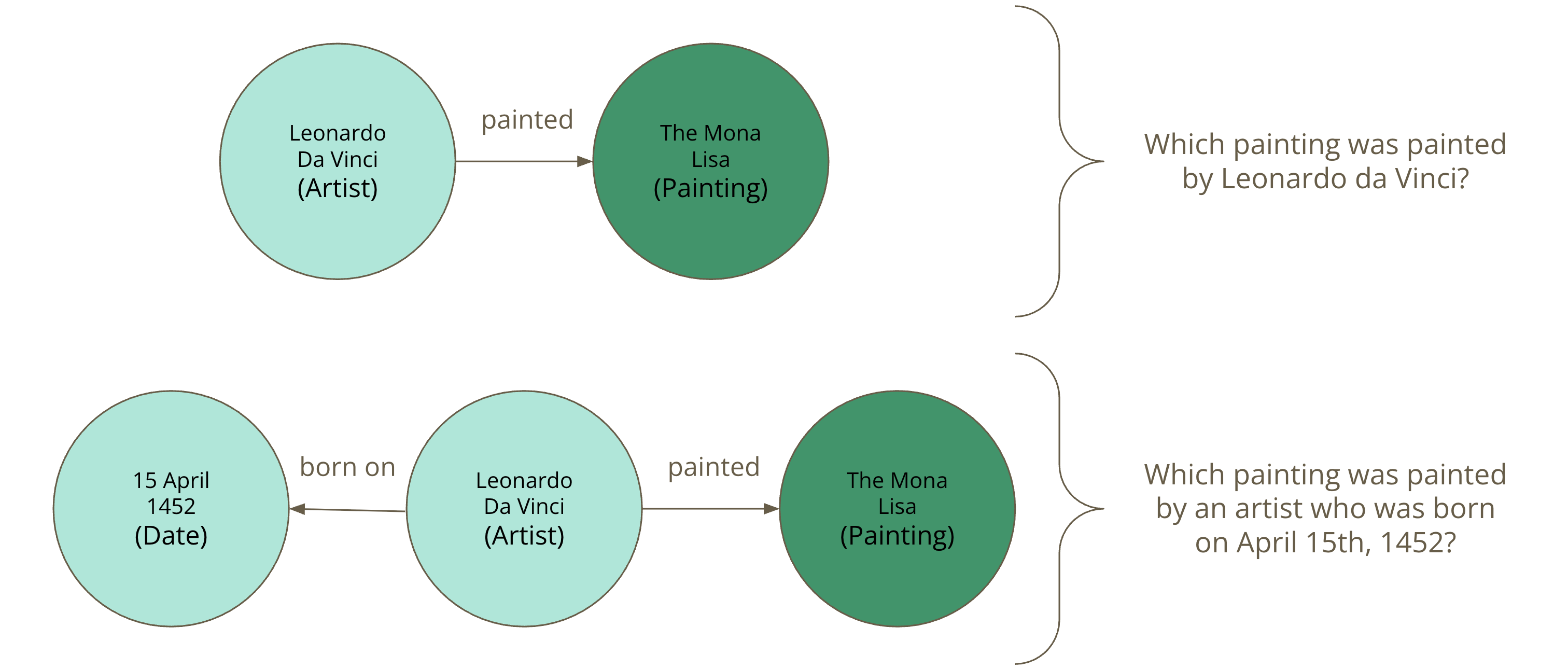}
\caption{Illustration of single-hop and double-hop reasoning for MCQ generation. The green node represents the key.}
\label{fig:signal_reasoning}
\end{figure}

\paragraph{\textbf{Reasoning}} This binary signal distinguishes between questions that are derived from a single triple (single-hop reasoning) and those constructed from a quintuple (double-hop reasoning). Double-hop questions are generally more cognitively demanding, as they require the integration of multiple relational facts. Figure~\ref{fig:signal_reasoning} illustrates this distinction.

\begin{figure}[htb]
\centering
\includegraphics[width=0.93\linewidth]{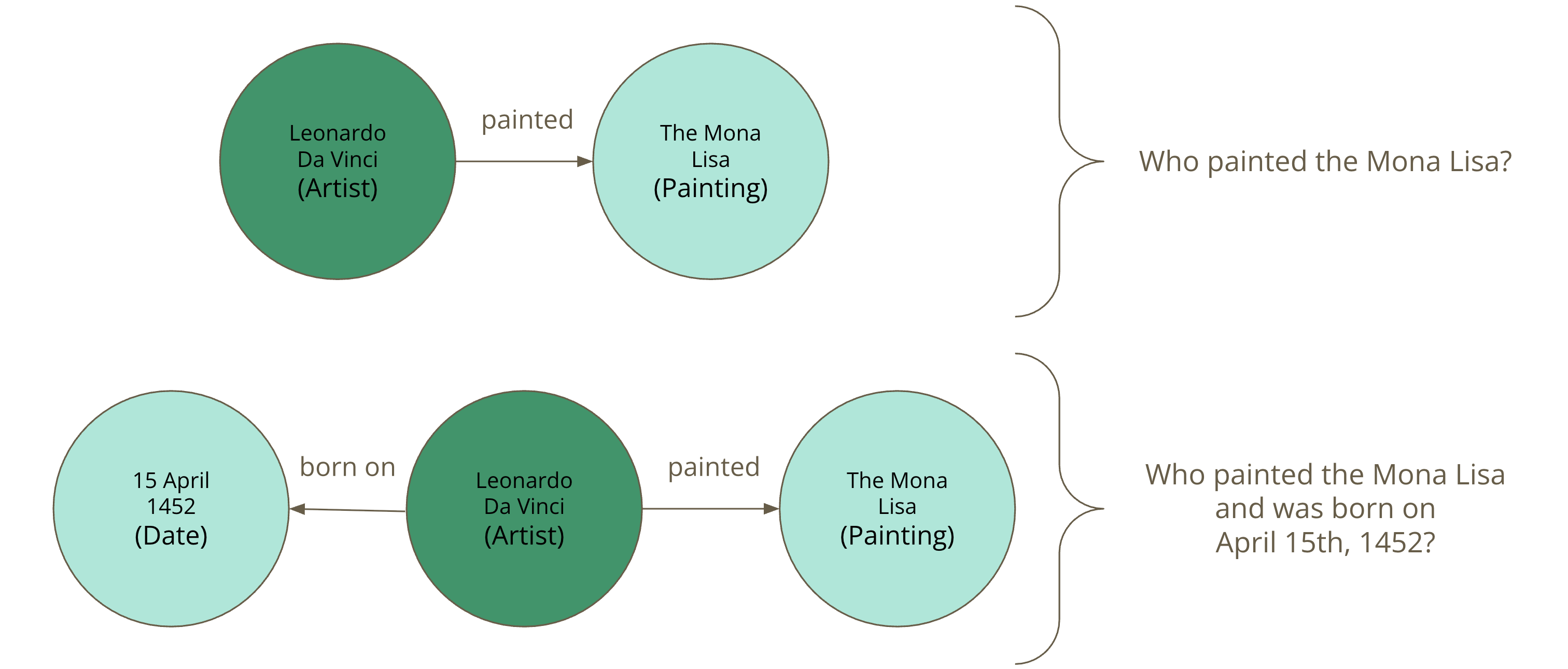}
\caption{Illustration of MCQ generation with and without an extra triple. The green node represents the key.}
\label{fig:signal_extratriple}
\end{figure}

\paragraph{\textbf{Extra Triple}} This binary signal captures whether an additional supporting triple is incorporated, providing further context. The inclusion of an extra triple may enrich the given knowledge inside the MCQ, but also potentially increase its complexity. An example is shown in Figure~\ref{fig:signal_extratriple}.

\begin{figure}[htb]
\centering
\includegraphics[width=0.93\linewidth]{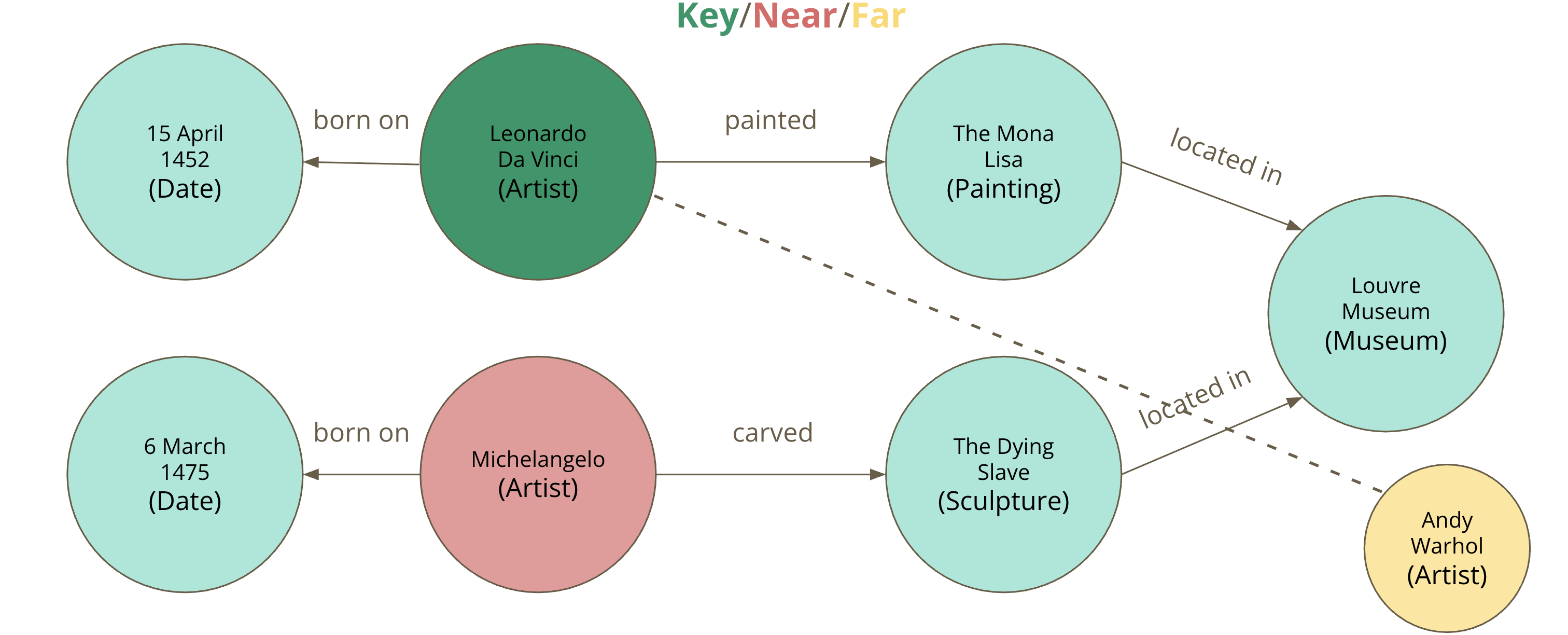}
\caption{Visualization of candidate distractor depths. The green node represents the key.}
\label{fig:signal_distractordepth}
\end{figure}

\paragraph{\textbf{Distractor Depth}} This signal measures the average depth of distractors from the key entity in the knowledge graph using breadth-first traversal. The farther the distractors are from the key entity, the less semantically related they tend to be, which typically reduces the difficulty. This structural notion is visualized in Figure~\ref{fig:signal_distractordepth}.

\paragraph{\textbf{Node Embedding Similarity}} We compute the average cosine similarity between the node embeddings of the distractors and the key, using FastRP embeddings, denoted as $e_n$. This is expressed in~\eqref{eq:signal_nodeembedsim}.

\begin{equation}
\text{\texttt{SignalNodeEmbedSim}} = \frac{1}{3} \sum_{i=1}^{3} \cos\left(e_n(\text{node}(d_i)), e_n(\text{node}(key))\right)
\label{eq:signal_nodeembedsim}
\end{equation}

\noindent Higher similarity implies more plausible distractors, increasing the MCQ's difficulty.

\paragraph{\textbf{Text Embedding Similarity}} This signal captures the semantic similarity between each distractor and the question stem in the embedding space. To contextualize the score, it is normalized by the similarity between the key and the stem. The computation is defined in~\eqref{eq:signal_textembedsim}.

\begin{equation}
\text{\texttt{SignalTextEmbedSim}} = 
\frac{\frac{1}{3} \sum_{i=1}^{3} \cos\left(e_t(\text{text}(d_i)), e_t(\text{stem})\right)}
     {\cos\left(e_t(\text{text}(key)), e_t(\text{stem})\right)}
\label{eq:signal_textembedsim}
\end{equation}

\noindent Here, $e_t$ denotes the text embedding function, specifically OpenAI’s \textsf{text-embedding-3-large}\footnote{\url{https://platform.openai.com/docs/models/text-embedding-3-large}} model.

\paragraph{\textbf{Degree Centrality}} The degree centrality of a node $v$ is defined as the number of incoming and outgoing edges it has. The formal definition is given in~\eqref{eq:signal_degreecentrality}.

\begin{equation}
\text{\texttt{SignalDegreeCentrality}} = \frac{1}{|V_s|} \sum_{v \in V_s} \text{deg}(v)
\label{eq:signal_degreecentrality}
\end{equation}

\noindent The signal takes the average degree centrality of all nodes involved in the associated knowledge subgraph. Higher centrality often implies more well-known concepts.

\paragraph{\textbf{Readability}} We compute the Flesch Reading Ease score, denoted as \texttt{F}, for the question stem, following the formulation proposed by \citet{flesch1948}. This score is used as the readability signal to quantify linguistic complexity. The computation is provided in~\eqref{eq:signal_readability}.

\begin{equation}
\text{\texttt{F}} = 
206.835 - 1.015 \cdot 
\left( \frac{\#\ \text{words}}{\#\ \text{sentences}} \right)
- 84.6 \cdot 
\left( \frac{\#\ \text{syllables}}{\#\ \text{words}} \right)
\label{eq:signal_readability}
\end{equation}

\paragraph{\textbf{Above Largest Gap Count}} We compute the cosine similarity between the question stem and each of the four distractor options, and sort the resulting similarity scores in descending order: $\text{sim}_1 \geq \text{sim}_2 \geq \text{sim}_3 \geq \text{sim}_4$. We then calculate the differences between adjacent similarity values and identify the position of the largest gap, drawing inspiration from the Maximum Gap Thresholding technique proposed by \citet{soykokguvenir}. The signal is defined as the number of distractors that appear before the largest gap in the sorted list, as formalized in~\eqref{eq:signal_abovelargestgapcount}.

\begin{equation}
\text{\texttt{SignalAboveLargestGapCount}} = \arg\max_{i \in \{1, 2, 3\}} \left( \text{sim}_i - \text{sim}_{i+1} \right)
\label{eq:signal_abovelargestgapcount}
\end{equation}

\noindent This ordinal signal captures how many distractors are grouped more closely in semantic similarity to the stem, potentially influencing the difficulty of identifying the correct answer.

\paragraph{\textbf{LLM Extra Fact}} This binary signal indicates whether the generated question stem includes factual content not directly inferable from the associated knowledge subgraph. Such additions may be unintended hallucinations by the LLM and typically increase the difficulty or ambiguity of the question. This signal is automatically generated for each MCQ by prompting an LLM to compare the question stem and the associated knowledge subgraph.

\begin{table}[htbp]
  \caption{Overview of MCQ‐generation signals and their descriptions.}
  \label{tab:signals}
  \centering
  \footnotesize
  \begin{tabular}{@{}p{0.24\linewidth}p{0.71\linewidth}@{}}
    \toprule
    \textbf{Signal} & \textbf{Description} \\
    \midrule
     Reasoning &
      A binary indicator denoting whether the MCQ was derived from a triple or a quintuple. \\\hline

     Extra Triple &
      A binary indicator representing the use of an extra triple for additional context. \\\hline

     Distractor Depth &
      The average graph distance between the key and each distractor. \\\hline

     Node Embedding Similarity &
      The average cosine similarity between the key and distractors based on their graph node embeddings. \\\hline

     Text Embedding Similarity &
      The average cosine similarity between the distractors and the question stem, divided by the cosine similarity between the key and the stem. \\\hline

     Degree Centrality &
      The average degree centrality of the entities involved in the question stem, reflecting their graph-level prominence. \\\hline

     Readability &
      The Flesch Reading Ease score of the question stem, indicating its linguistic complexity. \\\hline

     Above Largest Gap Count &
      An ordinal indicator capturing the number of distractor options that precede the largest semantic gap when cosine similarities between the question stem and each distractor are sorted in descending order. \\\hline

     LLM Extra Fact &
      A binary flag indicating whether the question stem introduces factual content not explicitly present in the selected subgraph. \\
    \bottomrule
  \end{tabular}
\end{table}

All signals are normalized to the $[0,1]$ interval using min-max normalization to ensure scale compatibility. Together, they serve as interpretable and complementary indicators of MCQ difficulty. Moving on, we explore how these signals are leveraged to train a supervised model for estimating difficulty scores. Table~\ref{tab:signals} summarizes the nine proposed signals.

\subsubsection{Difficulty Estimation Model}
After computing the defined difficulty signals for each validated MCQ, we train a supervised regression model to estimate the MCQ difficulty. The target variable is the empirical difficulty score, approximated by the observed incorrect response rate collected from users. This score lies within the $[0,1]$ interval, where higher values correspond to more difficult questions.

The proposed difficulty signals are model-agnostic and can be straightforwardly used as input features in any regression framework. We evaluate several standard regression models, specifically \textsf{LinearRegression}, \textsf{RandomForestRegressor}, and \textsf{GradientBoostingRegressor} from the \textsf{scikit-learn}\footnote{\url{https://scikit-learn.org/}} library, as well as \textsf{XGBRegressor} from the \textsf{XGBoost}\footnote{\url{https://xgboost.readthedocs.io/}} library. Among these, the \textsf{XGBRegressor} yields the best performance in our experiments. All models are trained using an 80/20 split, and we report evaluation metrics including root mean squared error (RMSE), mean squared error (MSE), mean absolute error (MAE), and coefficient of determination ($R^2$ score) for each.

The resulting regression model provides a continuous-valued difficulty score for each MCQ, which we refer to as its \textit{estimated difficulty}. In addition to enabling quantitative difficulty estimation, this approach allows us to interpret the relative importance of each difficulty signal through feature importance analysis.

\section{Dataset}
\label{section4}
To the best of our knowledge, there exists no available dataset that jointly provides (i) the KG or subgraph used to construct a multiple-choice question (MCQ), (ii) the MCQ itself, and (iii) a corresponding difficulty label. Given that our difficulty estimation framework relies on both the structural and semantic characteristics of the source knowledge graph as well as the inherent properties of the MCQ itself, we constructed a dedicated dataset for this task.

We began by collecting textual data from Wikipedia’s top-100 most popular articles, which we automatically retrieved using a custom scraping script.
These documents were then processed using LangChain’s \textsf{LLMGraphTransformer} with OpenAI’s GPT-4o model to construct a knowledge graph. The resulting KG captures semantic relationships and entity types, and serves as the foundation for MCQ generation.

To generate MCQs, we selected the top-40 most central nodes in the graph using degree centrality and attempted to generate four MCQs per node using our methodology. This yielded a total of 156 MCQs. We presented these MCQs to a diverse pool of human participants, and for each MCQ, we received approximately 38 responses on average. This approach enhances the reliability of the ground truth difficulty labels by reducing subjectivity and averaging out individual biases, as each MCQ's ground truth label depends on the responses of approximately 38 participants.

For each MCQ, we computed the \textit{incorrect answer rate}, the proportion of participants who answered the question incorrectly, and used it as the empirical ground truth difficulty score. Additionally, participants were asked to rate how much they liked each question on a standardized scale. These ratings were averaged to yield a quality metric for each MCQ. The mean liking score across all MCQs was $66\%$, indicating a generally favourable reception. Also, a negative correlation of $-0.49$ was observed between average liking scores and difficulty, suggesting that participants tended to prefer easier questions.

\begin{figure}[htb]
\centering
\includegraphics[width=0.65\linewidth]{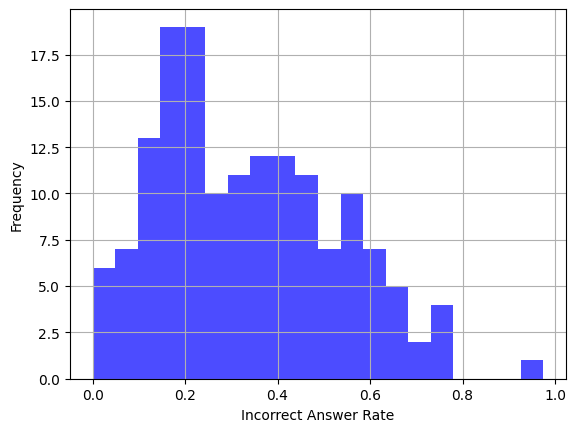}
\caption{Histogram of the incorrect answer rates of generated MCQs.}
\label{fig:difficulty_histogram}
\end{figure}

The histogram in Figure~\ref{fig:difficulty_histogram} shows the distribution of incorrect answer rates across the dataset. The mean incorrect response rate was $0.34$, indicating a moderate overall difficulty level. The distribution is slightly skewed toward easier questions, likely due to the intentional selection of high-centrality nodes in the knowledge graph, derived from Wikipedia’s most popular articles, to ensure familiarity and contextual relevance for participants.

This dataset offers a unique resource where each data point consists of the source knowledge subgraph, the generated MCQ, and an empirically derived difficulty label as the ground truth for the difficulty score. It enables modeling the difficulty using both KG–based and MCQ–based signal sources, effectively addressing our initial requirement. 

\section{Experimental Results and Analysis}
\label{section5}

This section presents the experimental evaluation of our proposed difficulty estimation framework. All experiments were conducted on a MacBook Pro equipped with an Apple M3 chip, integrated GPU, and 16~GB of unified memory. 
The system was running macOS with ARM64 architecture. 

We assess the effectiveness of multiple regression models using standard performance metrics and provide an analysis of predictive accuracy and feature importance.

\subsection{Evaluation Metrics}

To evaluate the performance of the regression models, we report the following metrics:
\begin{itemize}
\item \textbf{Root Mean Squared Error (RMSE):} Measures the square root of the average squared differences between predicted and actual values. RMSE penalizes larger errors more than smaller ones.
\item \textbf{Mean Absolute Error (MAE):} Represents the average absolute difference between predicted and actual values, offering a more interpretable error measure.
\item \textbf{Coefficient of Determination ($R^2$):} Indicates the proportion of variance in the actual difficulty scores explained by the predicted scores.
\item \textbf{Spearman's Rank Correlation ($\rho$):} Measures the strength and direction of the monotonic relationship between predicted and actual difficulty scores. This metric is particularly useful for evaluating how well the predicted rankings of MCQs align with the ground-truth rankings, thus reflecting the model’s ability to preserve the relative ordering in terms of MCQ difficulty.
\end{itemize}

These metrics collectively provide a comprehensive view of model accuracy, robustness, and consistency in ranking, which are important for assessing the quality of difficulty predictions.

\subsection{Model Selection and Results}

Given the limited size of our dataset, we restrict our analysis to classical regression models that are well-suited for small-to-moderate-sized datasets. Specifically, we evaluate \textsf{LinearRegression}, \textsf{RandomForestRegressor}, \textsf{GradientBoostingRegressor}, and \textsf{XGBRegressor}, leveraging \textsf{scikit-learn} and \textsf{XGBoost} libraries for implementation. Table~\ref{tab:model_comparison} summarizes the performance of these models.

\begin{table}[htbp]
\caption{Performance comparison of regression models on MCQ difficulty estimation. (The target variable, MCQ difficulty, is a continuous value in the range $[0,1]$.)}
\label{tab:model_comparison}
\centering
\begin{tabular}{lcccc}
\toprule
\textbf{Model} & \textbf{RMSE} & \textbf{MAE} & \boldmath{$R^2$} & \textbf{Spearman's $\boldsymbol{\rho}$} \\
\midrule
Linear Regression            & 0.14 & 0.12 & 0.46 & 62.4\% \\
Random Forest Regressor      & 0.14 & 0.12 & 0.41 & 64.2\% \\
Gradient Boosting Regressor  & 0.15 & 0.13 & 0.34 & 46.4\% \\
XGBoost Regressor            & \textbf{0.13} & \textbf{0.11} & \textbf{0.52} & \textbf{64.3\%} \\
\bottomrule
\end{tabular}
\end{table}

During our analysis, we identified an outlier MCQ instance in the dataset with an empirically derived difficulty score of $0.975$, the highest in the dataset and visibly higher than all others, as can be seen in Figure~\ref{fig:difficulty_histogram}. This particular instance was found to contain two semantically valid answer choices, which were not detected by either our KG-based or LLM-based validation mechanisms during generation. Moreover, the distractor labeled as incorrect appeared more plausible to users than the intended correct answer, leading to unusually high confusion and, consequently, an out-of-distribution empirical difficulty score. Although this is a single instance, we conducted our experiments both including and excluding it (the code provided under \textsf{Data availability} reproduces both variants), and the results demonstrated that excluding this faulty instance had minimal impact on the overall model performances, as expected given it is only one data point. The results reported in this paper are based on the version of the dataset with this MCQ excluded.

\subsection{Prediction Accuracy Visualization}

\begin{figure}[htb]
\centering
\subfloat[Linear Regression]{
    \centering
    \includegraphics[width=0.45\linewidth]{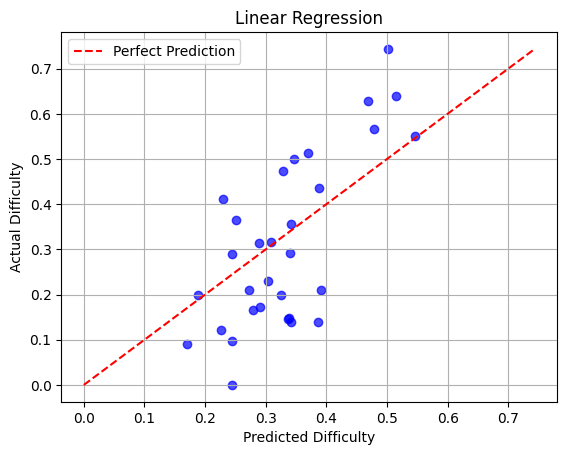}
    \label{fig:linear}
}
\hfill
\subfloat[Random Forest Regressor]{
    \centering
    \includegraphics[width=0.45\linewidth]{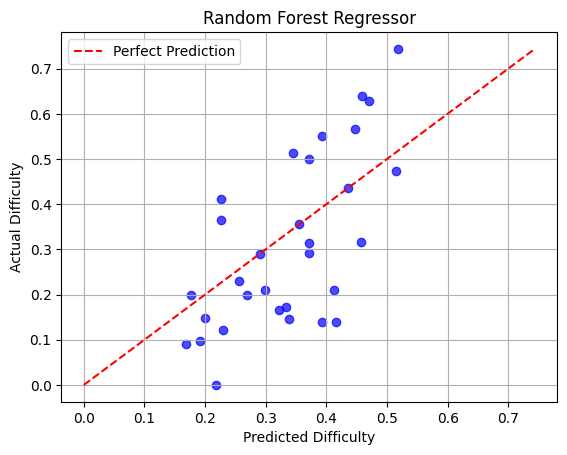}
    \label{fig:rf}
}
\vspace{0.05cm}

\subfloat[Gradient Boosting Regressor]{
    \centering
    \includegraphics[width=0.45\linewidth]{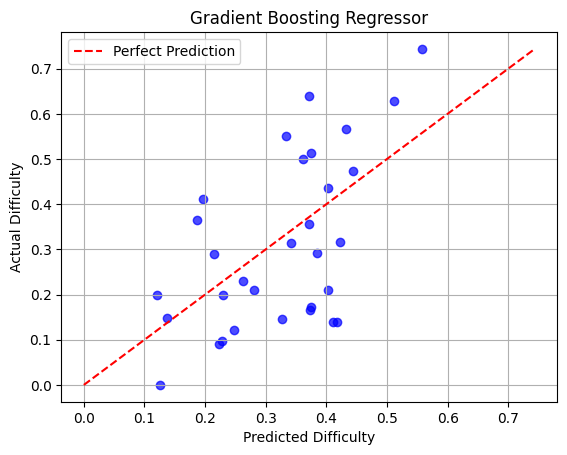}
    \label{fig:gb}
}
\hfill
\subfloat[XGBoost Regressor]{
    \centering
    \includegraphics[width=0.45\linewidth]{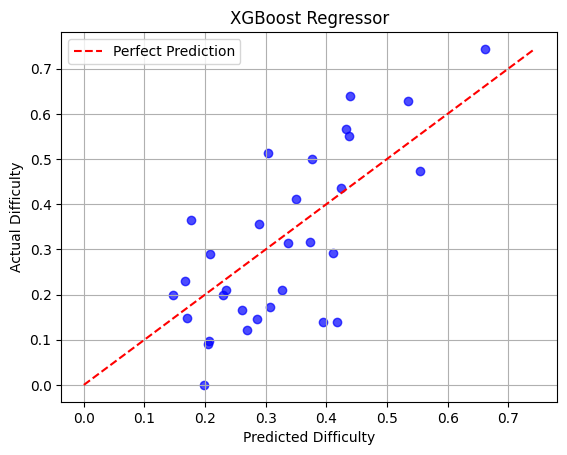}
    \label{fig:xgb}
}

\caption{Predicted vs actual difficulty for each regression model.}
\label{fig:pred_vs_actual}
\end{figure}

Figures~\ref{fig:linear}, \ref{fig:rf}, \ref{fig:gb}, and \ref{fig:xgb} visualize the predicted vs. actual difficulty scores for each regression model. The red dashed line represents the ideal case where predicted and actual values are equal.

\subsection{Feature Importance Analysis}

\begin{figure}[htb]
\centering
\includegraphics[width=0.85\linewidth]{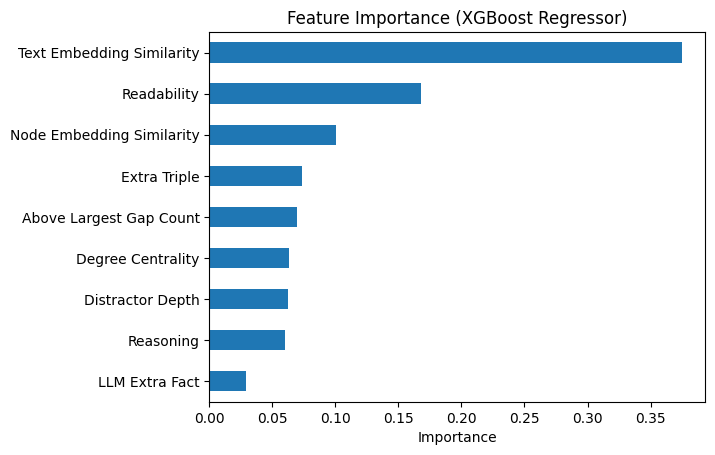}
\caption{Feature importance scores from XGBoost Regressor.}
\label{fig:xgb_feature_importance}
\end{figure}

To gain insight into which signals contribute most to the difficulty prediction, we examine the feature importances learned by the XGBoost Regressor, which yielded the best performance. The feature importance scores shown in Figure~\ref{fig:xgb_feature_importance} are derived from the \textsf{feature\_importances\_} attribute of the trained \textsf{XGBoost Regressor}. This metric quantifies the contribution of each input feature to the model’s predictions by measuring the average gain in performance (the amount of reduction in loss) when a feature is used for splitting across all trees in the ensemble. Features with higher scores are those that, on average, contribute more to accurate predictions.
The results are presented in Figure~\ref{fig:xgb_feature_importance}.

\begin{figure}[htb]
\centering
\includegraphics[width=0.85\linewidth]{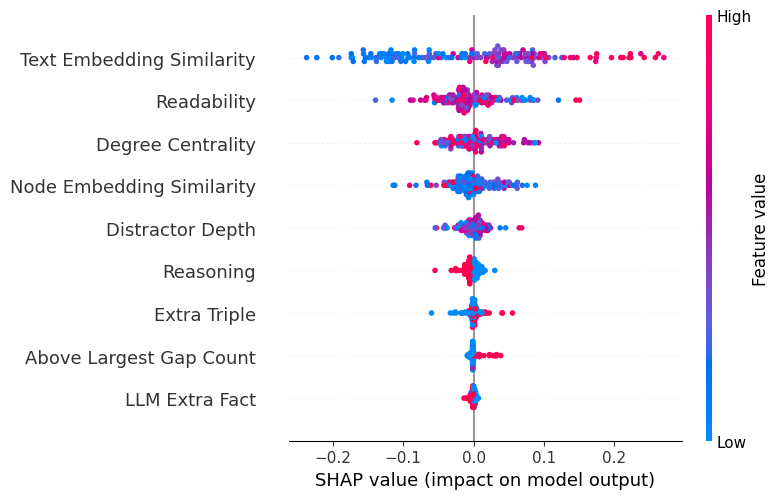}
\caption{SHAP summary plot for the XGBoost Regressor. Each point corresponds to a SHAP value for a feature and an instance.}
\label{fig:shap_summary_plot}
\end{figure}

To further interpret the model predictions, we use SHAP (SHapley Additive exPlanations) values~\citep{shap}. We compute SHAP values for the trained \textsf{XGBoost Regressor} using the method proposed by \citet{shap-tree}, and visualize the results using a summary plot, as shown in Figure~\ref{fig:shap_summary_plot}. The plot shows both the magnitude and direction of each feature’s impact across instances. Red points represent higher feature values, and blue points represent lower values, enabling a detailed visual of how individual features affect the model's output across varying input instances.
As seen in Figure~\ref{fig:shap_summary_plot}, the top two features and the last feature remain the same as in Figure~\ref{fig:xgb_feature_importance}.
Only minor changes are observed in the intermediate rankings, which indicates an overall similarity.

\subsection{Ablation Study}

\begin{figure}[htb]
\centering
\includegraphics[width=\linewidth]{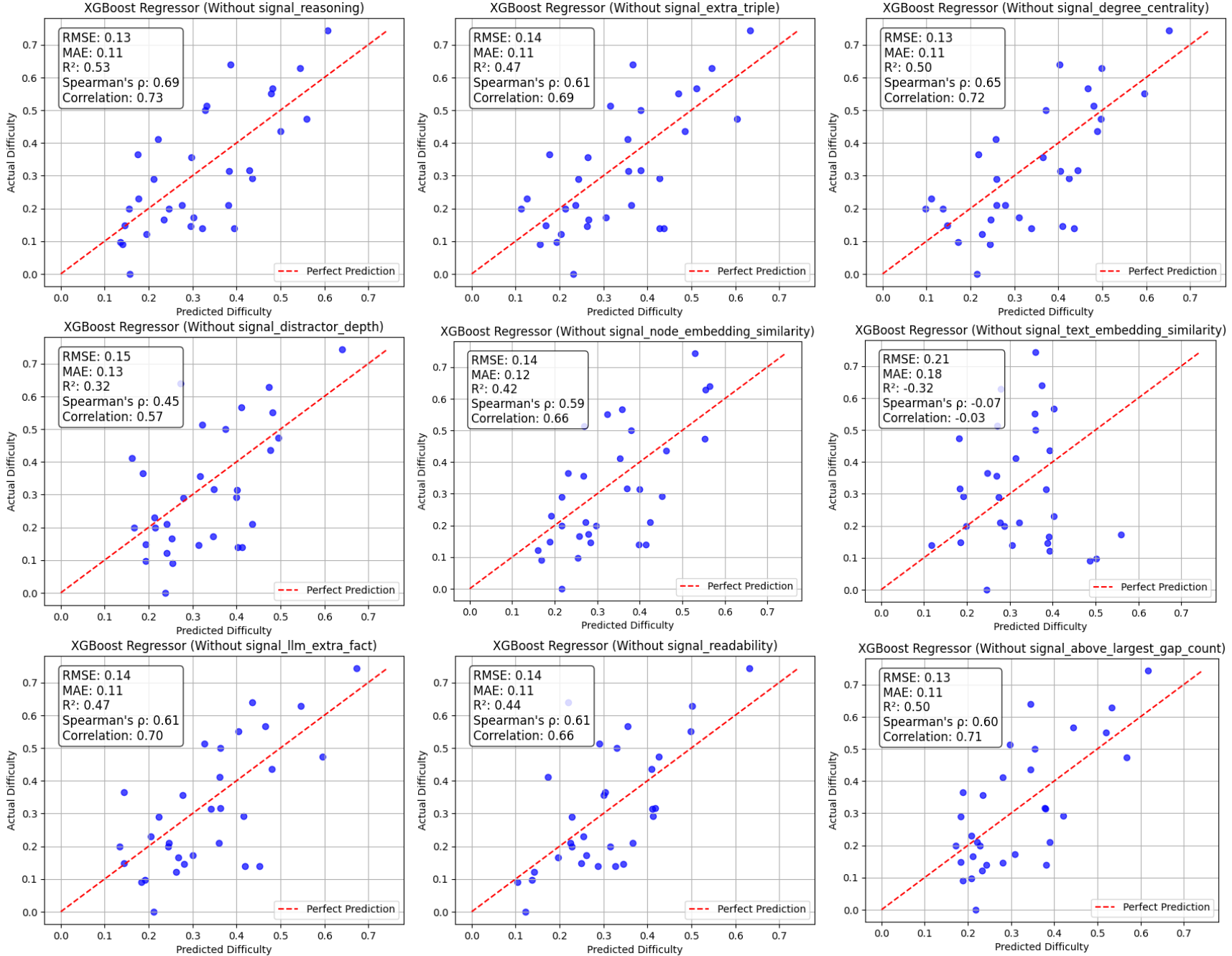}
\caption{Single-feature ablation study results.}
\label{fig:ablation_single}
\end{figure}

To evaluate the contribution of each individual signal to the difficulty estimation task, we conducted an ablation study in which we trained the \textsf{XGBoost Regressor}, our best-performing model, after excluding one signal at a time. The results are presented in Figure~\ref{fig:ablation_single}, where each plot shows the predicted vs.\ actual difficulty scores along with the metrics for the model trained without a specific feature.

We observed that the absence of three particular signals (\textsf{Reasoning}, \textsf{Degree Centrality}, and \textsf{Above Largest Gap Count}) did not degrade performance compared to our original best-performing model that used all the signals (RMSE = 0.13, MAE = 0.11). This observation motivated a second round of ablation experiments, in which we trained models excluding pairwise combinations of these three features.

\begin{figure}[htb]
\centering
\subfloat[2-feature ablation experiment.]{
    \centering
    \includegraphics[width=0.45\linewidth]{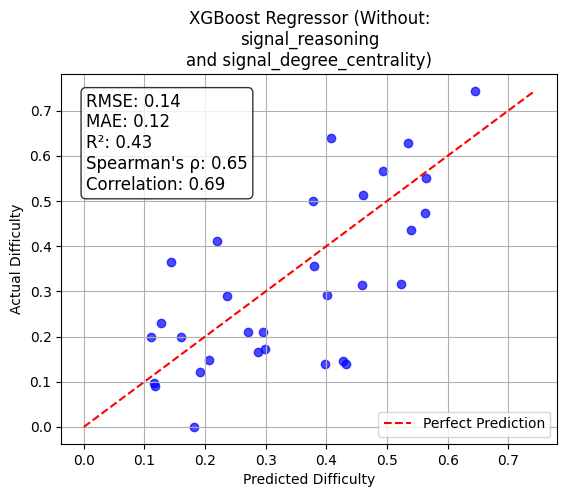}
    \label{fig:ablation_two_1}
}
\hspace{0.01\linewidth}
\subfloat[2-feature ablation experiment.\textbf{*}]{
    \centering
    \includegraphics[width=0.45\linewidth]{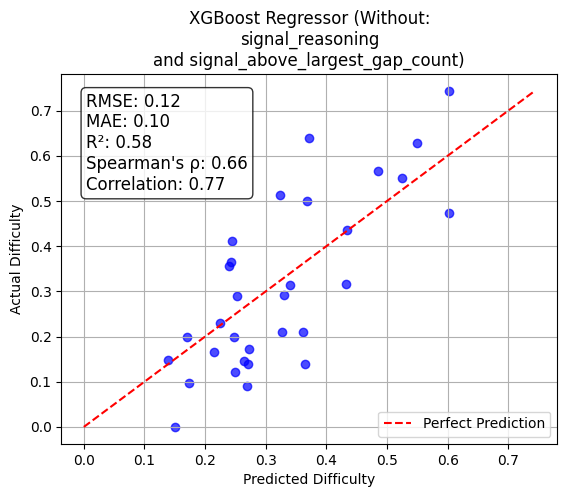}
    \label{fig:ablation_two_2}
}

\vspace{0.01cm}

\subfloat[2-feature ablation experiment.]{
    \centering
    \includegraphics[width=0.45\linewidth]{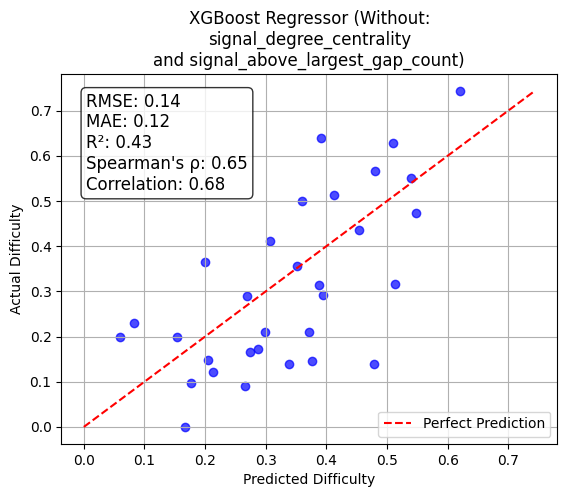}
    \label{fig:ablation_two_3}
}
\hspace{0.01\linewidth}
\subfloat[3-feature ablation experiment.]{
    \centering
    \includegraphics[width=0.45\linewidth]{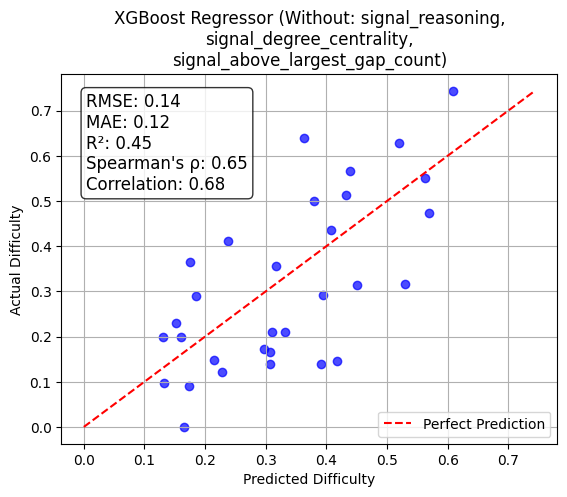}
    \label{fig:ablation_three}
}

\caption{Extended ablation study results for 2-feature and 3-feature exclusions.\newline
(a) Excluded: \textit{Reasoning} and \textit{DegreeCentrality}. No improvement.\newline
(b) Excluded: \textit{Reasoning} and \textit{AboveLargestGapCount}. \textbf{*Best-performing exp}.\newline
(c) Excluded: \textit{DegreeCentrality} and \textit{AboveLargestGapCount}. No improvement.\newline
(d) Excluded: \textit{Reasoning}, \textit{DegreeCentrality}, and \textit{AboveLargestGapCount}. No improvement.}
\label{fig:ablation_extended}
\end{figure}

Among the three possible 2-feature exclusion combinations, the configuration excluding \textsf{Reasoning} and \textsf{Above Largest Gap Count} achieved improved performance, with RMSE = 0.12 and MAE = 0.10, thereby surpassing our previous best-performing model. The other two combinations produced RMSE = 0.14 and MAE = 0.12, which were inferior to the baseline. For completeness, we also excluded all three features simultaneously, resulting in RMSE = 0.14 and MAE = 0.12, again worse than the baseline. The results of these experiments are presented in Figure~\ref{fig:ablation_extended}. Moreover, the results of the new best-performing model, compared with the previous best, are shown in Table~\ref{tab:ablation_best}.

\begin{table}[htbp]
\caption{Comparison of baseline and improved XGBoost Regressor configurations.}
\label{tab:ablation_best}
\centering
\begin{tabular}{lcccc}
\toprule
\textbf{Model} & \textbf{RMSE} & \textbf{MAE} & \boldmath{$R^2$} & \textbf{Spearman's $\boldsymbol{\rho}$} \\
\midrule
XGBRegressor - All signals & 0.13 & 0.11 & 0.52 & 64.3\% \\
XGBRegressor - Best & \textbf{0.12} & \textbf{0.10} & \textbf{0.58} & \textbf{66.1\%} \\
\bottomrule
\end{tabular}
\end{table}

\newpage

\section{Discussion}
\label{section6}

Our experiments show that all evaluated regression models produce reasonably strong performance given the size of the dataset, with the \textsf{XGBoost Regressor} consistently outperforming the others across all evaluation metrics. The $R^2$ score of $0.52$ and Spearman’s $\rho$ of $64.3\%$ for the baseline XGBoost configuration indicate a strong correspondence between predicted and actual difficulty scores, confirming the viability of our proposed difficulty estimation framework.

The ablation study results provide deeper insights into the role of individual signals. The absence of certain signals (\textsf{Reasoning}, \textsf{Degree Centrality}, and \textsf{Above Largest Gap Count}) did not reduce performance compared to the full-feature baseline. More importantly, removing one specific pair of features (\textsf{Reasoning} and \textsf{Above Largest Gap Count}) led to a new best-performing model with $RMSE = 0.12$, $MAE = 0.10$, $R^2 = 0.58$, Spearman's $\rho = 64.3\%$ and surpassing the original configuration.

Furthermore, the feature importance and SHAP analyses offer interpretable evidence regarding the contributions of individual signals. The observed results indicate that semantic features play a central role in difficulty estimation. Among them, \textsf{Text Embedding Similarity} emerges as the most influential signal, highlighting the critical role of semantic alignment between the distractors and the question stem, with respect to the alignment between the key and the stem. Notably, \textsf{Readability}, a relatively simple and long-established linguistic measure, also proves to be highly effective, reaffirming its continued relevance. Graph-based features such as \textsf{Node Embedding Similarity} also demonstrate meaningful contributions, supporting the utility of incorporating structural information from the underlying knowledge graph into difficulty estimation. In contrast, \textsf{LLM Extra Fact} exhibits relatively minimal influence, suggesting that its inclusion serves more as a precautionary indicator than a key predictive signal.

These findings demonstrate the viability of our approach for performing interpretable and accurate difficulty estimation on the MCQs generated from the input textual sources through our framework, using features grounded in both the underlying knowledge graph and the MCQs themselves.

Beyond the methodological contributions and discussions, our findings have practical implications for multiple application domains. In adaptive learning systems, automated MCQ generation with accurate and interpretable difficulty estimation can enable the dynamic selection of questions tailored to a learner’s proficiency, thereby improving engagement and learning outcomes.

Moreover, while the experiments were conducted on a dataset of limited size and domain scope, the underlying methodology is not inherently tied to any specific subject area. The combination of semantic, structural, and linguistic signals, together with a systematic feature evaluation process, makes the framework adaptable to other domains where multiple-choice knowledge questions are used or applicable. The same approach could be applied to different areas. This adaptability, together with the strong performance demonstrated in our experiments, underlines the potential for broader applicability beyond the specific setting examined in this work.

This study demonstrates that interpretable, high-performing difficulty estimation for automatically generated MCQs is achievable through a combination of semantic, structural, and linguistic features, with careful attention to feature evaluation and selection. The proposed framework offers a foundation for practical deployment in real-world educational and assessment contexts, and it opens multiple pathways for future research.

\section{Conclusion and Future Work}
\label{section7}

This study introduces a novel framework for generating multiple-choice knowledge questions with interpretable difficulty estimation by integrating knowledge graphs (KGs), which provide structured representations of factual knowledge, with large language models (LLMs), which offer advanced natural language understanding and generation capabilities. Starting from unstructured textual sources, we construct a KG that captures factual relationships, generate MCQs from selected subgraphs, and compute a suite of interpretable signals that reflect both structural and linguistic aspects of each MCQ. These signals are then used as input features to train regression models for predicting empirical difficulty scores, demonstrating reasonably strong alignment with human evaluations.

Our experimental results confirm the viability of this approach, showing that difficulty can be effectively estimated using features derived from both the underlying KG and the MCQs themselves. Among the evaluated models, XGBoost achieved the best performance across all metrics, and feature importance analysis further supported the value of combining multiple types of signals.

One key limitation of our work lies in the size of the collected dataset. Due to practical constraints, the number of MCQs and corresponding human responses was relatively limited. This restricted our ability to further explore more complex models requiring large volumes of training data, such as deep neural architectures. Future work could aim to scale data collection to support the training and evaluation of different and more sophisticated models.

Furthermore, while the current dataset primarily targets common knowledge, expanding it to incorporate specialized knowledge domains represents a promising direction for future work.
Additionally, exploring user modeling and personalized difficulty estimation offers another valuable direction for advancement.

Moreover, although failures in detecting invalid question generation attempts through either the KG-based or LLM-based validation mechanisms are currently rare, with only one such instance occurring in this work, future work could incorporate cross-model validation with multiple models to further reduce the likelihood of such cases.

Overall, our work establishes a comprehensive framework that generates multiple-choice knowledge questions given textual inputs and estimates their difficulty in an interpretable and accurate manner. By leveraging large language models for language understanding and generation capabilities and knowledge graphs for structured representation, our approach transforms raw texts into high-quality MCQs and provides interpretable difficulty estimation grounded in both semantic (graph-based and text-based) and linguistic characteristics. This end-to-end capability also opens new pathways for research at the intersection of knowledge representation, educational technology, and natural language understanding and generation. This approach has strong potential to support personalized learning and adaptive assessment in educational settings by enabling the generation of high-quality MCQs with interpretable difficulty levels.

\vspace{6pt} 





\noindent\paragraph{\bf{Author contributions:}:} Conceptualization, H.A.G. and K.K.; methodology, M.C.Ş.; software, M.C.Ş.; validation, M.C.Ş. and H.A.G.; data curation, M.C.Ş.; writing---original draft preparation, M.C.Ş., H.A.G., and K.K; writing---review and editing,  M.C.Ş., H.A.G., and K.K; visualization,  M.C.Ş., H.A.G., and K.K; supervision, H.A.G.; project administration, H.A.G. and K.K.; funding acquisition, H.A.G. and K.K. All authors have read and agreed to the published version of the manuscript.

\noindent\paragraph{\bf{Funding}:} This research was supported by the European Union under the Horizon Europe Research and Innovation Action (RIA) project titled Customized Games and Routes for Cultural Heritage and Arts (Grant Agreement No. 101094428). The authors gratefully acknowledge the funding provided by the European Commission, which enabled the development of this study. The views and opinions expressed are those of the authors and do not necessarily reflect those of the European Union or the granting authority.

\noindent\paragraph{\bf{Institutional Review}:} This study, related to our dataset collection experiment, has been approved by the Bilkent University IRB with approval number 662 on 01/03/2025.



\noindent\paragraph{\bf{{Abbreviations}}:} The following abbreviations are used in this manuscript:
\\

\noindent 
\begin{tabular}{@{}ll}
MCQ & Multiple-choice question\\
KG & Knowledge graph\\
LLM & Large language model\\
RNN & Recurrent neural network\\
GNN & Graph neural network\\
GAT & Graph attention network\\
SPARQL & SPARQL Protocol and RDF Query Language\\
NER & Named entity recognition\\
BFS & Breadth-first search\\
MoE & Mixture of experts\\
BiLSTM & Bidirectional long short-term memory\\
RTRL & Relation-aware transformer with reinforcement learning\\
KGNN-ADP & Knowledge graph neural network-based adaptive difficulty prediction\\
AutoDE & Automatic difficulty estimation\\
FastRP & Fast random projection\\
RMSE & Root mean squared error\\
MSE & Mean squared error\\
MAE & Mean absolute error\\
R\textsuperscript{2} & Coefficient of determination\\
SHAP & SHapley Additive exPlanations\\
\end{tabular}

\bibliography{bibliography}

\end{document}